\documentclass{article}

\usepackage{arxiv}

\usepackage[utf8]{inputenc} 
\usepackage[T1]{fontenc}    
\usepackage{hyperref}       
\usepackage{url}            
\usepackage{booktabs}       
\usepackage{amsfonts}       
\usepackage{nicefrac}       
\usepackage{microtype}      
\usepackage{lipsum}
\usepackage{graphicx}
\graphicspath{ {./images/} }

\usepackage{listings}
\title{Comparative Analysis of Drug-GPT\textsuperscript{\texttrademark} and ChatGPT LLMs for Healthcare Insights: Evaluating Accuracy and Relevance in Patient and HCP Contexts}

\author{
Giorgos Lysandrou\\
george@talkingmedicines.com
\and
Roma English Owen\\
roma@talkingmedicines.com
\and
Kirsty Mursec\\
kirsty@talkingmedicines.com
\and
Grant Le Brun\\
grant@talkingmedicines.com
\and
Elizabeth A. L. Fairley $\ast$\\
elizabeth@talkingmedicines.com\\
$\ast$ corresponding author
}

\begin{document}
\maketitle
\begin{abstract}

This study presents a comparative analysis of three Generative Pre-trained Transformer (GPT) solutions in a question and answer (Q\&A) setting: Drug-GPT\textsuperscript{\texttrademark} 3, Drug-GPT\textsuperscript{\texttrademark} 4, and ChatGPT, in the context of healthcare applications. The objective is to determine which model delivers the most accurate and relevant information in response to prompts related to patient experiences with atopic dermatitis (AD) and healthcare professional (HCP) discussions about diabetes. The results demonstrate that while all three models are capable of generating relevant and accurate responses, Drug-GPT\textsuperscript{\texttrademark} 3 and Drug-GPT\textsuperscript{\texttrademark} 4, which are supported by curated datasets of patient and HCP social media and message board posts, provide more targeted and in-depth insights. ChatGPT, a more general-purpose model, generates broader and more general responses, which may be valuable for readers seeking a high-level understanding of the topics but may lack the depth and personal insights found in the answers generated by the specialized Drug-GPT\textsuperscript{\texttrademark} models. This comparative analysis highlights the importance of considering the language model's perspective, depth of knowledge, and currency when evaluating the usefulness of generated information in healthcare applications.

\end{abstract}


\section*{Introduction}
Generative Pre-trained Transformers (GPTs) are state-of-the-art large language models (LLMs) that have garnered significant attention in recent years, particularly with the introduction of ChatGPT, due to their ability to generate human-like text. These models have demonstrated great potential in various natural language processing (NLP) tasks, such as text classification, question answering and summarisation, making them highly relevant for applications across diverse domains, including healthcare.

Healthcare companies have acknowledged the enormous potential of LLMs in revolutionizing their operations. By harnessing the capabilities of LLMs, these organizations can extract valuable insights, enhance decision-making processes, and ultimately improve patient outcomes. Some notable applications of LLMs in healthcare include:

\begin{itemize}
    \item \textsc{Supporting Clinical Decision Making:} LLMs can examine extensive medical literature and clinical guidelines to offer evidence-based recommendations for HCPs, aiding in diagnosis, treatment planning, and personalized care \cite{rao2023evaluating}.
    \item \textsc{Facilitating Drug Discovery and Development:} LLMs contribute to the identification of patterns and potential drug candidates by analysing molecular data, scientific literature, and clinical trials. This expedites the drug discovery process and empowers researchers to make well-informed choices \cite{sharma2023chatgpt}.
     \item \textsc{Aiding Medical Translation:} LLM's can enable translation that aids in effective communication between healthcare providers and patients. Leveraging advanced language processing capabilities, LLM's can both swiftly and precisely translate medical and technical terminology into everyday language. This can help patients' understanding of their diagnosis and treatment options \cite{Marr_2023}.
\end{itemize}




In this comparative analysis, we evaluate the performance of three GPT solutions in a question and answer (Q\&A) setting: Drug-GPT\textsuperscript{\texttrademark} 3 and Drug-GPT\textsuperscript{\texttrademark} 4, which are based on GPT-3 and GPT-4 respectively, and supported by curated datasets of patient and HCP social media and message board posts, and ChatGPT, a more general-purpose model, to assess their performance in the healthcare domain. Our objective is to determine which model delivers the most accurate and relevant information, demonstrated by responses to prompts related to patient experiences with AD and HCP discussions about diabetes. By comparing the capabilities of these models, we aim to highlight their strengths and limitations and provide insights into their potential applications in healthcare.

\section*{Background }
Li et al. (2023) conducted a systematic review of existing publications on the use of ChatGPT in healthcare, aiming to provide insights into the current state of ChatGPT in medical applications for general readers, HCPs, and NLP scientists. The authors found that the current release of ChatGPT has achieved only moderate performance in various tests and is unreliable for actual clinical deployment, as it is not designed for clinical applications. They concluded that specialized NLP models trained on biomedical datasets still represent the right direction for critical clinical applications. While this work highlights the importance of utilising specialised data, our approach opts to include this data through prompt engineering instead of training, which has the benefits of retaining the LLM's general natural language generation abilities, while still delivering specialised insights on real-world patient and HCP experiences. The review also discusses ethical concerns and potential pitfalls of using ChatGPT in healthcare, 
emphasizing the need for creators and providers of chatbots like ChatGPT to address these issues to avoid catastrophic consequences \cite{li2023chatgpt}.

Salvagno et al. (2023) discuss the potential use of artificial intelligence (AI) chatbots, specifically ChatGPT, in scientific writing. The authors suggest that ChatGPT can assist researchers in organizing material, generating initial drafts, and proofreading. However, they emphasize that AI-generated work should not replace human judgment and should always be reviewed by experts, which is why Drug-GPT\textsuperscript{\texttrademark} is aimed at providing healthcare insights to aid, not replace, human decision-making. The paper also raises ethical concerns, such as the risk of plagiarism, inaccuracies, and potential imbalances in accessibility between high and low-income countries due to costs \cite{salvagno2023can}.

Cascella et al. (2023) explored the potential applications and limitations of ChatGPT, in healthcare. Although ChatGPT has shown promising results in various NLP tasks and even successfully passed the United States Medical Licensing Exam (USMLE), its performance in real-world clinical scenarios remains uncertain. The authors investigated the feasibility of ChatGPT in four clinical and research scenarios: supporting clinical practice, scientific production, potential misuse in medicine and research, and reasoning about public health topics. While ChatGPT demonstrated an ability to generate coherent and realistic text, the study highlights the importance of recognizing and promoting education on the appropriate use and potential pitfalls of AI-based LLMs in medicine, such as "hallucinated" answers, a behaviour we take cautious measures to avoid with Drug-GPT\textsuperscript{\texttrademark}. Additionally, ethical concerns surrounding the use of ChatGPT in scientific articles warrant further attention \cite{cascella2023evaluating}.

In a study conducted by Johnson et al. (2023), the authors assessed the accuracy and reliability of ChatGPT-generated medical responses. The study involved 33 physicians across 17 specialities, generating 284 medical questions. These questions were then answered by ChatGPT, and the physicians graded the responses for accuracy and completeness using specified Likert scales. The results demonstrated that ChatGPT generated largely accurate information for diverse medical queries as judged by academic physician specialists, with a median accuracy score of 5.5 (between almost completely and completely correct) and mean score of 4.8 (between mostly and almost completely correct). However, important limitations were also identified, indicating the need for further research and model development to correct inaccuracies and ensure validation. This study provides an early evidence base for the potential of AI-based systems like ChatGPT, or even more specialised healthcare solutions such as Drug-GPT\textsuperscript{\texttrademark}, in providing accurate and comprehensive medical information, while also emphasizing the need for cautious implementation in medical practice and research \cite{johnson2023assessing}.

Wang et al. (2023) introduce ClinicalGPT, an LLM specifically designed and optimized for clinical scenarios. Recognizing the limitations of existing LLMs in medical applications, the authors leverage extensive and diverse real-world data, including medical records, domain-specific knowledge, and multi-round dialogue consultations, to train ClinicalGPT. A comprehensive reinforcement learning evaluation framework is also introduced, encompassing medical knowledge question-answering, medical exams, patient consultations, and diagnostic analysis of medical records. The results reveal that ClinicalGPT significantly outperforms other models in these tasks, demonstrating the effectiveness of their approach in adapting LLMs to the critical domain of healthcare. This study underscores the importance of incorporating domain-specific knowledge and diverse data sources, like the approach we took with Drug-GPT\textsuperscript{\texttrademark} and its specialised curated patient and HCP dataset, to enhance the utility of LLMs in specialized fields such as medicine \cite{wang2023clinicalgpt}.

\section*{Methodology}

The focus of these experiment comparisons is on GPT language models, developed by OpenAI, leveraging unsupervised learning and large-scale pre-training to generate human-like text by predicting the next word in a sentence given its context. The first GPT model, GPT-1 \cite{radford2018improving}, laid the foundation for its successors, followed by GPT-2 \cite{radford2019language}, GPT-3 \cite{brown2020language} and GPT-4 \cite{openai2023gpt4}, which offer increasingly sophisticated linguistic capabilities. GPT-3 was a big breakthrough in performance, and hence it was released to the public as part of a chat system called ChatGPT.

ChatGPT, a chatbot application initially backed by GPT-3, was designed specifically for generating contextually relevant and meaningful responses in a conversational setting. While it shares the core architecture and pre-training approach with GPT-3, ChatGPT incorporates essential modifications to GPT-3 to better cater to the unique requirements of chatbot applications. To enhance its conversational capabilities, GPT-3 is fine-tuned on custom datasets that contain conversational data, enabling it to learn the nuances of human dialogue and generate context-aware responses, resulting in ChatGPT. Additionally, the model is subjected to Reinforcement Learning from Human Feedback (RLHF) to iteratively improve its performance by learning from user interactions. This allows ChatGPT to stand out from traditional GPT models, providing a more effective solution for generating human-like responses in conversational AI applications.

Drug-GPT\textsuperscript{\texttrademark} is an advanced and specialized proprietary tool not available for public use, unlike ChatGPT, aimed at offering valuable insights regarding patients and HCPs, in a Q\&A setting.
Based on empirical research evidence demonstrating that specialized LLMs outperform general ones in specific domains like healthcare \cite{li2023chatgpt}\cite{johnson2023assessing}\cite{wang2023clinicalgpt}, we have opted for the approach of utilizing a specialized LLM solution to ensure better accuracy and reliability in our targeted application. This solution utilizes a carefully curated patient and HCP dataset derived from social media and message board posts, a sophisticated information retrieval system, and the GPT-3 or GPT-4 LLMs. Owing to the distinct LLMs supporting Drug-GPT\textsuperscript{\texttrademark} 3 and Drug-GPT\textsuperscript{\texttrademark} 4, the latter is capable of accommodating a larger input, hence considering a larger context from its knowledge base before generating a response. By implementing specifically tailored hyperparameters and prompt engineering, Drug-GPT\textsuperscript{\texttrademark} delivers reproducible responses based on the most relevant content within its knowledge base dataset. The provided answers are factual, stemming from real-world experiences of patients and HCPs, rather than being derived from the LLM's own knowledge. Consequently, Drug-GPT\textsuperscript{\texttrademark}'s knowledge encompasses a broader scope compared to GPT-3, GPT-4, or ChatGPT.

The comparison is based on the Q\&A answers provided by Drug-GPT\textsuperscript{\texttrademark} 3, Drug-GPT\textsuperscript{\texttrademark} 4 and ChatGPT in response to the following prompts:

\begin{enumerate}
    \item Experiment 1: \textsc{Patients}: “What are three challenges of living with atopic dermatitis?”.
    \item Experiment 2: \textsc{HCPs}: “What are the top themes HCPs are talking about in relation to diabetes?”.
\end{enumerate}

The choice of hyperparameters plays a crucial role in obtaining reproducible outcomes. For this purpose, we set the Temperature parameter to 0, thereby ensuring that the model generates the most confident answers without any randomness. Moreover, we set the Top P parameter to 1, which enables the model to select the most probable words during the token sampling process. This approach facilitates a deterministic and consistent output for a given question. These selection criteria not only enhance the overall coherence and accuracy of the generated text, through avoiding random, diverse and less probable responses, but also provide a reliable basis for evaluating the models' efficacy in the context of the given tasks, because the generated response is always the most probable.

The experiments were conducted using the hyperparameters specified in Table \ref{tab:hyperparameters}. The hyperparameters of the general-use ChatGPT were not made publicly available by OpenAI, but the version of ChatGPT that produced our results is the February 13th 2023 release. The results are analysed in terms of relevance and accuracy.

\begin{table}[h]
\centering
\begin{tabular}{llll}
\hline
Name              & \textbf{ChatGPT} & \textbf{GPT-3}     & \textbf{GPT-4}     \\ \cline{2-4} 
Model name        & -                & text-davinci-003   & gpt-4 8k           \\
Version           & Feb 13 2023 release   & 2022-12-01 release & 2023-03-15-preview \\
Max tokens        & -                & 512                & 2048               \\
Temperature       & -                & 0                  & 0                  \\
Top P             & -                & 1                  & 1                  \\
Frequency penalty & -                & 0                  & 0                  \\
Presence penalty  & -                & 0                  & 0                  \\ \hline
\end{tabular}
\caption{Experiment General Pre-trained Transformer Large Language Models hyperparameters detailed setup.}
\label{tab:hyperparameters}
\end{table}

\section*{Experiments}

In these experiments, the models were given a prompt, with the models' responses being analysed to determine their accuracy, relevance, and conciseness. The goal was to identify any commonalities in the responses and assess the depth and quality of information provided by each solution.

\subsection*{Experiment 1: Patients}

The comparison is based on the answers provided by Drug-GPT\textsuperscript{\texttrademark} 3, Drug-GPT\textsuperscript{\texttrademark} 4, and ChatGPT in response to the following prompt:

\textit{"What are three challenges of living with atopic dermatitis?"}

What the question asks for can be answered the best by first-hand experience of patients experiencing AD. While ChatGPT only has access to the general data it was pre-trained and fine-tuned with, Drug-GPT\textsuperscript{\texttrademark} 3 and 4 have access to a curated patient dataset, with patients specifically experiencing AD. Based on the question, Drug-GPT\textsuperscript{\texttrademark} 3 and 4 can provide an answer by retrieving the most relevant patient experiences, talking expressly about challenges of living with AD.

\subsection*{Experiment 2: HCPs}

The comparison is based on the answers provided by Drug-GPT\textsuperscript{\texttrademark} 3, Drug-GPT\textsuperscript{\texttrademark} 4, and ChatGPT in response to the following prompt:

\textit{"What are the top themes HCPs are talking about in relation to Diabetes? Provide the answer as a list highlighting the context with a quote from the context."}

The question asks for information around HCP discussions about diabetes, which can be best answered with real-world HCP discussions in this domain. ChatGPT again only has access to the general information found in its pre-training and fine-tuning data, while Drug-GPT\textsuperscript{\texttrademark} 3 and 4 have access to a curated HCP dataset, of online HCP interactions. Drug-GPT\textsuperscript{\texttrademark} 3 and 4 based their answer on the retrieved conversations around diabetes.

\section*{Results}

The complete prompts for the questions and answers provided by the LLMs in the following experiments, can be seen in the appendix section Prompts \& Answers.

\subsection*{Experiment 1: Patients}

All three models provided answers that highlighted the challenges of living with AD. The full prompt questions and answers can be seen in section "Experiment 1: Patients", in "Prompts \& Answers" in the appendix. A summary of each model's response is as follows:

\begin{enumerate}
    \item Drug-GPT\textsuperscript{\texttrademark} 3:
    \begin{enumerate}
        \item Damage caused by long-term use of topical steroids.
        \item Overwhelming nature of finding symptom solutions.
        \item Difficulties managing food allergies and sensitivities.
    \end{enumerate}
    \item Drug-GPT\textsuperscript{\texttrademark} 4:
    \begin{enumerate}
        \item Damage from long-term use of topical steroids.
        \item Identifying and managing food triggers and allergies.
        \item Finding effective treatments and managing flare-ups.
    \end{enumerate}
    \item ChatGPT:
    \begin{enumerate}
        \item Skin irritation and discomfort.
        \item Social isolation and psychological impact.
        \item Treatment difficulties.
    \end{enumerate}
\end{enumerate}

All three models agreed that finding treatment options for AD is a challenge. ChatGPT provided a longer but still relevant response, which included an introduction to the health condition. The responses from Drug-GPT\textsuperscript{\texttrademark} 3 and Drug-GPT\textsuperscript{\texttrademark} 4 were more concise but still relevant. All three models could identify significant challenges faced by individuals with AD, with ChatGPT providing a more verbose answer. Drug-GPT\textsuperscript{\texttrademark} 3 and Drug-GPT\textsuperscript{\texttrademark} 4's answers focus on specific aspects, such as the damage caused by prescribed topical steroids and the difficulty in managing food triggers and allergies. These answers demonstrate a more targeted approach and seem to address issues that might be faced by patients in their treatment process.

The answer provided by ChatGPT, is more general and could be perceived as coming from someone who does not have first-hand experience with the condition but possesses knowledge about it. The response outlines three broader challenges: skin irritation and discomfort, social isolation and psychological impact, and treatment difficulties. The answer provides a topline overview of the condition, which may be valuable for readers looking for a general understanding of AD. However, this general approach might lack the depth and personal insights that could be found in answers generated by models with a more focused, first-hand understanding of the condition and its treatments. 

Drug-GPT\textsuperscript{\texttrademark}'s responses were noticeably more informal and accessible to laypeople in nature. Accessing patient experiences expressed in the way patients themselves discuss them is of utmost importance for the industry to improve comprehension and address the issues patients face. This distinction is attributed to Drug-GPT\textsuperscript{\texttrademark}'s restriction of generation of answers from a curated collection of patient social media posts, contrasting with the broader range of data sources that ChatGPT was trained on. Overall, the comparison highlights the importance of considering the language model's perspective, the information it considers in order to generate a response, and depth of knowledge in assessing the usefulness of generated information.

\subsection*{Experiment 2: HCPs}

All three models provided answers that highlighted the top themes discussed by HCPs in relation to diabetes. The full prompt questions and answers can be seen in section "Experiment 2: HCPs", in "Prompts \& Answers" in the appendix. A summary of each model's response is as follows:

\begin{enumerate}
    \item Drug-GPT\textsuperscript{\texttrademark} 3:
    \begin{enumerate}
        \item Sustained diabetes control with thiazolidinediones (TZD).
        \item Prevention of progression from prediabetes to diabetes with acarbose.
        \item Adjusting diabetes regimens for better control.
        \item Various drug treatments for diabetes and related conditions.
        \item Risk of pancreatitis with certain diabetes drugs.
        \item First line treatment for type 1 diabetes: multiple daily injections or continuous insulin infusion.
        \item Postprandial hypoglycemia and potential causes.
        \item Complications of diabetes medications.
        \item Explanation of type 1 diabetes: autoimmune attack on insulin-producing cells.
        \item Misdiagnosis of diabetes and its rarity.
    \end{enumerate}
    \item Drug-GPT\textsuperscript{\texttrademark} 4:
    \begin{enumerate}
        \item Diabetes control and A1c improvement with thiazolidinediones (TZD).
        \item Medications for diabetes: metformin and empagliflozin.
        \item Acarbose lowers progression from prediabetes to diabetes.
        \item Adjusting diabetes regimen through primary care doctor.
        \item Treatment for type 1 diabetes: prandial and basal insulin or continuous subcutaneous insulin infusion.
        \item Postprandial hypoglycemia due to increased insulin sensitivity.
        \item Misdiagnosis of diabetes and its potential causes.
    \end{enumerate}
    \item ChatGPT:
    \begin{enumerate}
        \item Prevention of diabetes.
        \item Treatment options for diabetes.
        \item Individualized care for diabetic patients.
        \item Lifestyle modifications in diabetes management.
        \item Technology in diabetes care and management.
    \end{enumerate}
\end{enumerate}


Drug-GPT\textsuperscript{\texttrademark} 3 and Drug-GPT\textsuperscript{\texttrademark} 4 provided relevant themes, with both of them including quotes for each theme. ChatGPT provided a list with five themes and accompanying quotes, which were accurate and relevant. The results showed that all three models were able to capture the top themes in diabetes-related discussions among HCPs. Drug-GPT\textsuperscript{\texttrademark} 3's answer includes quotes from various sources, addressing specific themes such as disease progression prevention, drug side effects and risks, and disease misdiagnosis. Drug-GPT\textsuperscript{\texttrademark} 4's response, is another comprehensive list of similar, equally specific themes and context. Both of these answers seem to be more focused on the medical aspects of diabetes and its complications.

ChatGPT's answer appears to be more general and could be perceived as coming from someone who is not themselves an HCP, but has a basic understanding of the topic. The response provides a list of themes along with quotes from the context, covering aspects such as prevention, treatment options, individualized care, lifestyle modifications, and technology. With respect to currency, it is important to note that ChatGPT's answers are outdated by 2 or 3 years, as seen in the publication dates provided alongside its sources, which may affect the relevance and accuracy of the information provided. While the answer offers a broad overview of the themes discussed by HCPs, it may lack the depth, specificity, and up-to-date knowledge that could be found in answers generated by models with a more focused understanding of the medical aspects of diabetes.

When comparing the answers provided by ChatGPT and Drug-GPT\textsuperscript{\texttrademark}, it was evident that Drug-GPT\textsuperscript{\texttrademark}'s responses exhibited a significantly higher similarity to the content and level of detail found in patient support materials and prescribing information commonly used within the industry \cite{Center_for_Drug_Evaluation_and_Research}. This comparison also highlights the importance of considering the language model's perspective, the information it considers in order to generate a response, depth of knowledge, and currency of the response when evaluating the usefulness of the generated information.

\section*{Discussion}

The comparative analysis of GPT language models in this study revealed that specialized models like Drug-GPT\textsuperscript{\texttrademark} 3 and Drug-GPT\textsuperscript{\texttrademark} 4, which leverage curated datasets of patient and HCP experiences, demonstrate a more focused approach in generating accurate and contextually relevant insights. In contrast, ChatGPT, a general-purpose model, was able to provide coherent and factually correct responses, but could lack the depth and specificity offered by the specialized models.

In Experiment 1, Drug-GPT\textsuperscript{\texttrademark} 3 and Drug-GPT\textsuperscript{\texttrademark} 4 can provide more targeted and specific answers, in part to their access to curated patient datasets. They were able to identify challenges faced by individuals with AD, focusing on aspects such as the damage caused by prescribed topical steroids and the difficulty in managing food triggers and allergies. ChatGPT, on the other hand, provided a more general response that, while still relevant, may lack the depth and personal insights that could be found in the answers generated by models with a more focused, first-hand understanding of the condition and its treatments.

In Experiment 2, Drug-GPT\textsuperscript{\texttrademark} 3 and Drug-GPT\textsuperscript{\texttrademark} 4 again demonstrated more focused understanding, providing relevant themes supported by quotes for context for each theme. ChatGPT's answer was more general, covering broader themes such as prevention, treatment options, individualized care, lifestyle modifications, and technology. However, ChatGPT's answers were outdated by 2 or 3 years, which may affect the relevance and accuracy of the information provided.

The experiments showcased the ability of Drug-GPT\textsuperscript{\texttrademark} 3 and Drug-GPT\textsuperscript{\texttrademark} 4 to retrieve real-world experiences and insights from patients and HCPs, illustrating their potential in delivering more targeted and valuable information. The answers generated by these specialized models were concise, relevant, and encompassed a broader scope of knowledge compared to ChatGPT.

While ChatGPT provided comprehensive and general answers, its responses lacked the first-hand experiences and insights that are central to understanding the nuances of the conditions and treatments discussed. Moreover, the responses generated by ChatGPT appeared to be slightly outdated, which might affect the relevance of the information provided. This observation emphasizes the importance of using specialized models in domain-specific applications, as they can provide more up-to-date, accurate, and contextually rich information.

In both experiments, the choice of hyperparameters played a crucial role in obtaining consistent and reproducible outcomes. The deterministic approach adopted by setting the Temperature parameter to 0 and the Top P parameter to 1 ensured the generation of coherent and accurate text while providing a reliable basis for evaluating the models' efficacy. In comparison, with the Temperature parameter set to anything above 0 and Top P set to anything below 1, we sacrifice repeatability and accuracy for creativity, which increases the risk of "hallucinated" answers \cite{beutel2023artificial}.


The results of this study indicate that specialized models like Drug-GPT\textsuperscript{\texttrademark} 3 and Drug-GPT\textsuperscript{\texttrademark} 4 have clear distinctions over general-purpose models like ChatGPT when addressing specific domain-related questions. Their focus on curated datasets, prompt engineering, and tailored hyperparameters enable them to deliver more accurate, reliable, and contextually relevant information. However, it is crucial to consider the language model's perspective, the information it considers in order to generate a response, depth of knowledge, and currency of the response when evaluating the usefulness of the generated information.



\section*{Conclusion}

In this comparative analysis, we evaluated the performance of three GPT solutions in a Q\&A setting: Drug-GPT\textsuperscript{\texttrademark} 3, Drug-GPT\textsuperscript{\texttrademark} 4, and ChatGPT, in the context of healthcare applications. Our experiments focused on assessing the models' ability to provide accurate and relevant information in response to prompts related to patient experiences with AD and HCP discussions about diabetes.

The results demonstrated that all three models were capable of generating relevant and accurate responses. However, Drug-GPT\textsuperscript{\texttrademark} 3 and Drug-GPT\textsuperscript{\texttrademark} 4, which were supported by a curated dataset of patient and HCP social media and message boards posts, provided more targeted and in-depth insights into the specific challenges faced by patients and the themes discussed by HCPs. ChatGPT, a more general-purpose model, generated broader and more general responses, which may be valuable for readers seeking a high-level understanding of the topics but may lack the depth and personal insights found in the answers generated by the specialized Drug-GPT\textsuperscript{\texttrademark} models.

Additionally, it is essential to address potential hallucinations of answers and ensure responsible use of these AI tools. Ethical concerns and potential dangers associated with AI-generated information should not be overlooked. Ensuring that AI-generated content is reviewed by experts and does not replace human judgment is crucial for responsible implementation in medical practice and research.

This comparative analysis highlights the importance of considering the language model's perspective, depth of knowledge, and currency when evaluating the usefulness of generated information in healthcare applications. While general-purpose models like ChatGPT can provide valuable insights, specialized models like Drug-GPT\textsuperscript{\texttrademark} 3 and Drug-GPT\textsuperscript{\texttrademark} 4, which are supported by domain-specific datasets, offer a more focused understanding of the medical aspects and real-world experiences of patients and HCPs. This distinction is crucial for healthcare organizations seeking to harness the potential of LLMs to improve patient care, enhance decision-making processes, and revolutionize their operations.

\bibliographystyle{unsrt}
\bibliography{references}  






\appendix

\section*{Author details}

Talking Medicines Limited, SC447227

25 Blythswood Square, Top Floor, G2 4BL, Glasgow, Scotland, UK

\section*{Acknowledgements}

We would like to express our gratitude to Beatrice Alex, Lewis Boyd, and Ellen Halliday, as well as the Talking Medicines Founders Jo-Anne Halliday, Scott F. Crae and Elizabeth A.L. Fairley for all of their support of this publication.

\section*{Prompts \& Answers}
\label{prompts_answers}

\subsection*{Experiment 1: Patients}

\subsubsection*{Prompt question:}

\begin{lstlisting}
What are 3 challenges of living with atopic dermatitis? 
\end{lstlisting}

\subsubsection*{Drug-GPT\textsuperscript{\texttrademark} 3 answer:}

\begin{lstlisting}
1. Damage caused by decades of prescribed topical steroids. 
2. Finding solutions for symptoms can be overwhelming. 
3. Food allergies and sensitivities can be difficult to manage.
\end{lstlisting}

\subsubsection*{Drug-GPT\textsuperscript{\texttrademark} 4 answer:}

\begin{lstlisting}
1. Damage caused by long-term use of prescribed topical steroids. 
2. Identifying and managing food triggers and allergies that can exacerbate the condition. 
3. Finding effective treatments and managing flare-ups, which can be overwhelming and frustrating.
\end{lstlisting}

\subsubsection*{ChatGPT answer:}

\begin{lstlisting}
Atopic dermatitis (AD), also known as eczema, is a chronic inflammatory skin condition that affects millions of people worldwide. Some of the challenges of living with atopic dermatitis include:
1. Skin irritation and discomfort: People with atopic dermatitis experience persistent itching, dryness, and redness on their skin. These symptoms can cause discomfort and even pain, which can be distracting and affect their daily activities and quality of life.
2. Social isolation and psychological impact: Atopic dermatitis can be visibly noticeable and often leads to social stigma and embarrassment, especially in more severe cases. As a result, people with AD may avoid social activities, feel self-conscious about their appearance, and even experience depression and anxiety.
3. Treatment difficulties: While there are various treatment options available for atopic dermatitis, finding the right treatment can be challenging. Some treatments may cause side effects or not work for everyone, and managing the condition can require a lot of time and effort, which can be frustrating and exhausting for some individuals.
\end{lstlisting}

\subsection*{Experiment 2: HCPs}

\subsubsection*{Prompt question:}

\begin{lstlisting}
What are the top themes HCPs are talking about in relation to Diabetes? Provide the answer as a list highlighting the context with a quote from the context.
\end{lstlisting}

\subsubsection*{Drug-GPT\textsuperscript{\texttrademark} 3 answer:}

\begin{lstlisting}
    1. Diabetes control that is sustained - "Excellent diabetes control that is sustained. Depending on the population we're talking about, typical A1c improvement with a TZD can be up to 1.5%. This is comparable to metformin and GLP1a, significantly more than most of the other new drugs. It is also comparable to SUs - but the difference there is it is *sustained*."
    2. Prevention of progression from prediabetes to diabetes - "There's OK evidence with acarbose lowering progression to diabetes - it lowers it by about 20-25%, but the problem is that almost no one can tolerate acarbose without severe bloating and flatulence. But there's absolutely zero evidence that glipizide will prevent progression from prediabetes to diabetes."
    3. Adjusting diabetes regimens - "I'm not necessarily second-guessing a provider who has his complete history, but it might be worth talking to his primary care doctor to see if his diabetes regimen needs to be adjusted. Especially considering that his diabetes is not well-controlled based on his A1C. Also looks like he might have borderline iron deficiency.."
    4. Drug treatments for diabetes - "to group them:  diabetes:  &gt;metformin   &gt;empagliflozin  blood pressure, kidney protection:  &gt;ramipiril   blood thinner, typically used for heart attack/stroke prevention in someone at particularly high risk (who has had an event before), such as someone with a prior history of tia  &gt;clopidogrel   prostate enlargement: &gt;tadalafil ..."
    5. Risk of pancreatitis with certain drugs - "Liraglutide, semaglutide, dulaglutide can be used at any egfr. The DPP4i have variable metabolism. Linagliptin doesn't need any dose adjustment for renal excretion for example. They do probably increase the risk of pancreatitis. Other notable risk includes angioedema."
    6. First line treatment for type 1 diabetes - "Here is the section on 1st line treatment copied directly from the guideline you linked to me:  "Most people with type 1 diabetes should be treated with multiple daily injections of prandial and basal insulin, or continuous subcutaneous insulin infusion.""
    7. Postprandial hypoglycemia - "Your body could just be more sensitive to glucose in your body, so when it sees a sugar load from a meal it may secrete more insulin than normal. There are other issues that can cause postprandial hypoglycemia (like after bariatric surgery, some autoimmune diseases, etc) but from what knowledge I have, it sounds like you don't have any of these."
    8. Complications of diabetes medications - "In the context of something like diabetes, this can be extra complicated because your body is responding to a lot of different signals, and that is complicated further when you're on a blood sugar lowering medication."
    9. Explanation of type 1 diabetes - "Type 1 diabetes is a disease where your body attacks the cells in the pancreas that produce a hormone called insulin. Insulin's function in the body is to regulate the uptake of glucose into the cells. Glucose is a simple sugar, and is the main source of energy for the body. Without insulin, no glucose can be used by the body."
    10. Misdiagnosis of diabetes - "There are people who truly are misdiagnosed - a single a1c 6.5 and never again, and someone coded them as DM, or someone who requested a refill of metformin for their PCOS and a covering provider associated it with DM, etc. but those are very few and far between."
\end{lstlisting}



\subsubsection*{Drug-GPT\textsuperscript{\texttrademark} 4 answer:}

\begin{lstlisting}
    1. Diabetes control and A1c improvement: "Depending on the population we're talking about, typical A1c improvement with a TZD can be up to 1.5%."
    2. Medications for diabetes: "diabetes:  &gt;metformin    &gt;empagliflozin"
    3. Progression from prediabetes to diabetes: "There's OK evidence with acarbose lowering progression to diabetes - it lowers it by about 20-25%"
    4. Adjusting diabetes regimen: "it might be worth talking to his primary care doctor to see if his diabetes regimen needs to be adjusted."
    5. Treatment for type 1 diabetes: "Most people with type 1 diabetes should be treated with multiple daily injections of prandial and basal insulin, or continuous subcutaneous insulin infusion."
    6. Postprandial hypoglycemia: "Your body could just be more sensitive to glucose in your body, so when it sees a sugar load from a meal it may secrete more insulin than normal."
    7. Misdiagnosis of diabetes: "There are people who truly are misdiagnosed - a single a1c 6.5 and never again, and someone coded them as DM, or someone who requested a refill of metformin for their PCOS and a covering provider associated it with DM, etc."
\end{lstlisting}

\subsubsection*{ChatGPT answer:}

\begin{lstlisting}
Sure, here's a list of the top themes that HCPs are talking about in relation to diabetes, along with a quote from the context to support each theme:

Prevention: "Prevention is key, and I think that's where we need to go with diabetes. I think we need to focus on prevention." (Source: Medscape Education Diabetes Patient Management: Continuous Glucose Monitoring and Insulin Management in Diabetes Management, Published June 2020)

Treatment options: "There are many different classes of diabetes medications available, and it's important to choose the right one for each patient based on their individual needs and preferences." (Source: American Diabetes Association Standards of Medical Care in Diabetes-2021, Published January 2021)

Individualized care: "Individualizing treatment regimens and empowering patients to self-manage their diabetes are critical for achieving optimal outcomes." (Source: American Diabetes Association Standards of Medical Care in Diabetes 2021, Published January 2021)

Lifestyle modifications: "Healthy lifestyle modifications, including diet and exercise, are important components of diabetes management and should be emphasized to all patients." (Source: American Diabetes Association Standards of Medical Care in Diabetes -2021, Published January 2021)

Technology: "Technology, such as continuous glucose monitoring and insulin pumps, can help patients with diabetes achieve better glycemic control and improve their quality of life." (Source: American Diabetes Association Standards of Medical Care in Diabetes -2021, Published January 2021)
\end{lstlisting}

\end{document}